\documentclass[]{article}
\date{10 November 2009}

\usepackage{url}
\usepackage{fullpage}

\begin{document}

\title{Publishing Identifiable Experiment Code And Configuration Is Important, Good and Easy} 

\author{Richard T.~Vaughan \hspace{1cm} Jens Wawerla \vspace{5mm} \\   
Simon Fraser University \vspace{5mm}\\  \tt \{vaughan,wawerla\}@sfu.ca}

\maketitle

\section{Introduction}

A few months ago, a graduate student in another country called me
(Vaughan) to ask for the source code of one of my multi-robot
simulation experiments. The student had an idea for a modification
that she thought would improve the system's performance. By the
standards of scientific practice this was a perfectly reasonable
request and I felt obliged to give it to her. With our original code,
the student could (i) re-run our experiments to verify that we
reported the results correctly; (ii) inspect the code to make sure
that it actually implements the algorithm described in our paper;
(iii) change parameters and initial conditions to make sure our
results were not a fluke of the particular experimental setting; (iv)
modify the robot controllers and quantitatively compare her new method
with our originals. It would cost me nothing to make her a copy of our
code, and her methodology would be impeccable. Why then do we read so
few papers using this methodology?

It turned out to be impossible to identify exactly which code was used
to perform the experiments in our years-old paper. We had not labeled
the source code at that moment, and it had subsequently been
modified. All the code was under version control, so we could obtain
approximately the right code by looking at revision dates. But having
only {\it approximately} the right code strictly invalidates the
replication of the experiments. The user has no way of knowing what
the differences are between the code she has and the code we used. So
we were able to offer the requesting student some code that may or may
not be that used in the paper. This was better than nothing, but not
good enough, and we suspect this is quite typical in our community.

This disappointing episode was a wake-up call for me, and our group
has been discussing how we can make sure this doesn't happen again.
We propose to {\it routinely} publish the exact code for each
experiment that we use to justify any claims at all.

This paper explains why we think complete experiment publication is
{\bf important} and why it is {\bf good} for the originating
researchers as well as subsequent users. The second half of the paper
examines current standards of scientific data and code publications,
and cites some evidence of its benefits. But first we present our 
protocol for publishing {\it identifiable code} and show how {\bf easy} it is to do.

\section{Publishing code is easy}

\subsection{What to do}

\begin{enumerate}
\item The complete and exact source code, build scripts, configuration
files, maps, log analysis scripts, list of critical external
dependencies, details of run-time environment and any other
instructions and resources necessary for a skilled researcher to
replicate the experiment should be packaged and placed in public for
free and anonymous download by any reader. 
\item The source package
  should be labeled with a textual identifier that is specific to the version
  described in the paper, and the paper should contain the
  identifier. 
\end{enumerate}

\subsection{Doing it easily}
This can be easily and cheaply achieved as follows. The code is
assembled into an archive file (tarball, gzip, etc), and a digital
signature is obtained using a cryptographic hash function such as
SHA1\cite{RFC3174}. The archive is published at some reliable
Internet host, and its URI and signature are published in the
paper. The archive can also be linked from the authors' web
publication list.

Upon downloading the file, the user can determine that the archive
matches the signature in the paper. Use of a good hash identifier
makes it very difficult for authors to modify the code by mistake or
design, without this being detectable by the user/reader.

Modern software development tools make for an even easier
process. Revision control systems like
Git\footnote{http://git-scm.com/} automatically generate a SHA1
cryptographic hash key for each committed version, such that there is
a low probability of any two packages or versions having the same
key. The entire revision control database can be easily cloned from an
URI, and users can check out the correct version by its signature,
while still having access to later versions. The differences between
versions are easy to inspect using Git's tools. We have chosen this
approach, and are hosting our Git repositories at the independent host
GitHub\footnote{http://github.com}. While GitHub's tools and
convenience are currently compelling, the ideal host would be a
reliable and long-lived independent institution such as a university,
national library or professional organization such as the IEEE.

The idea of identifying a code package with a crypotgraphic hash is
not new -- on the contrary the standard way to distribute packages on
the Internet such that the downloader can trust that they received the
``real thing'' is for the author to post a web page containing a URI
to an archive file along with the SHA1 hash of the same file. Our
contribution here is to recommend this practice to our community as a
means to achieve the benefits of trustworthy code publication. 
%


\section{Publishing code is important}

\subsection{Falsifiability and shared artifacts}
Publishing the actual experiment alongside the paper which describes
and interprets it increases the scientific and practical value of the
work. It goes a long way to solving a problem our field faces from a
philosophy of science point of view: the fact that we are a synthetic
science that creates and studies artifacts, rather than a natural
science that studies an extant universe common to all scientists. By
reproducing and sharing our artifacts we synthesize a common
environment. 

Scientific claims are required to be falsifiable. If I make a claim in
a paper about a system I created, and to which you do not have access,
my claim is not falsifiable in practice. My claims are more
scientifically valuable if I make them as easy to falsify as possible,
which I can achieve by publishing the artifacts.

\subsection{Repeatability and quantitative comparison}
In the natural sciences experimental results gain credibility after
they are independently repeated at least once. In order to be able to
repeat an experiment, we often require many details that are not
available in the paper. As we are often able to make the exact and
entire experiment available for replication at negligible cost, we can
achieve the best possible repeatability.

Of course, we can not prove experiments are correct and while simply
re-running a program is not a strong validation, even this alone can
show up mistakes. A stronger validation is obtained by completely
re-implementing the code, or the important parts of it, but by testing
the new version using the original setting, as determined by
inspecting the original code, we can improve our confidence in the
results.

As in all of science, much work in robotics can be considered
incremental improvement over the work of another. This usually
requires reimplementing the original experiment from natural language
and formal mathematical descriptions. This re--implementation step
usually allows only qualitative comparison, since the details of
parameters and initial conditions, etc, are rarely published. It can
also be a source of error and raises question such as ``did the new
author really find the very best parameter set?''. Experiments made
public in an executable form will improve fairness to the original
author and will allow quantitative comparison of results.

A second level of repeatability is available to us. Components of
experiments can be re-used in different experiments and settings. If
the component performs as expected in this new setting, our confidence
in it increases. In fact this re-use of code is a cheap way of
reproducing experiments. 

\section{Publishing code is good } 
\subsection{Efficiency}
\label{sec:eff1}
Having access to data sets and software implementations increases the
efficiency of the scientific process in several ways. In the case of
incremental work, it saves a great deal of re-implementation
effort. While the use of middleware like Player\cite{player2001},
ROS\footnote{http://www.ros.org}, and Microsoft Robotics Developer
Studio\footnote{http://msdn.microsoft.com/robotics} has
increased the rate of code reuse in recent years, these systems focus
on low-level components and it is still unusual for a robot controller
or an implementation of an algorithm to be substantially
reused. Making code available by default would encourage reuse,
particularly if the code is of good quality.


\subsection{Quality}
The quality of a research contribution is a function of the soundness
and originality of its theoretical foundation, the depth of analysis
given and the clarity and thoroughness of its presentation. It is
assumed that the software that produces the results is correct. Yet it
is all too easy to make implementation mistakes that grossly influence
the outcome of an experiment. Even when a paper presents a complete
formal algorithm, discrepancies between the description and the
implementation that produced the results are possible. Such
discrepancies are impossible to detect without access to the source
code. We can very easily make code available for peer review, and so
we should.

Further, it is often argued that well written and documented software
has fewer bugs. Developing software with the expectation that it will
be peer reviewed and reused is likely to cause roboticists to write
better code, thus increasing the overall quality of the work even
before external review. We should write code as we write papers: to be
read and understood; to contribute to knowledge.

\section{Issues and objections}

Achieving code publication requires a number of issues to be
addressed. Some of the most significant are:

\begin{enumerate}
\label{sec:eff2}
\item {\it ``I object! All that extra work takes too long....''}

  There are three arguments. First, while producing peer-reviewable
  code may {\it feel} like it takes longer, the additional discipline
  and code review should result in improved code quality. By reducing
  bug-hunting and re-runs of faulty experiments, the experimenter
  could actually save time compared to a typical messy code
  base. Second, starting with others' published code saves time in the
  first place. Third, extra work is justified by the methodological
  advantages: the main role of the ``extra'' work is to improve the
  quality and usefulness of the research results, thus it should not
  be considered overhead. 

  Some documentation is usually required for code to be
  usable. Extensive end-user documentation can be very costly to
  produce, but often just a few notes can be sufficient to guide a
  colleague through reproducing and understanding an
  experiment. Documentation, particularly when written in-line with
  the code, can also help the author's confidence in its
  correctness. This idea is well developed in Knuth's {\it literate
    programming} model\cite{knuth:literate}.


\item {\it Offline vs. online} For work that does not need to
  interactively control the robot (e.g. offline SLAM) logs of original
  sensor data along with the algorithm code allow perfect experiment
  reproduction. The Radish repository exists to curate such datasets,
  and some of its maps have become familiar
  \footnote{http://radish.sourceforge.net}. Sharing interactive robot
  code is more troublesome.

\item {\it Real world and unique robots:} For interactive controller
  experiments, complete reproduction may be straightforward when
  experiments are done in simulation only. Yet real-robot experiments
  are essential. The arguments for experiment source sharing still
  hold for real robot systems, and the value of the work is maximized
  if the authors facilitate replication and extension. This can be
  done by using a well-known robot e.g. Pioneer, Khepera, which can be
  assumed to be widely available in research labs around the
  world. Well-known robots also have the advantage that respected
  simulation models are readily available. 
  
  If a custom robot is essential, we suggest providing either (i) a
  model for a well-known simulator, or (ii) a dedicated simulator
  including source code. Also when using custom hardware, using a
  well-known and open API for controller code, (e.g. Player, ROS)
  makes porting to another robot or simulator as easy as currently
  possible. Ideally, in all cases where a simulation can produce
  similar results to the real robot with a reasonable amount of effort
  that simulation should be provided.

  Work that uses novel mechatronics will not be reproducible at the
  very low cost of software-only or common-robot systems, but we
  suggest that engineering drawings, CAD models, materials and
  machining details can be published using the same methodology. The
  goal is to minimize the cost of reproduction. Work that does not
  require novel mechatronics should use a well known hardware and
  software platform. Section \ref{sec:willow} below discusses a recent
  major effort to bootstrap a widely-used platform.

\item {\it Licensing:} The free reading, copying, modification and
  subsequent redistribution of modified code is absolutely
  required. In most jurisdictions copyright law automatically applies,
  so the code must be explicitly licensed to allow redistribution. The
  community already makes extensive use of Free and Open Source
  Software, so we have experience with suitable licenses. Licences
  that prohibit commercial uses may be acceptable for traditional
  academic purposes, but clearly make the code less valuable for some
  users (the difference in value being what they are willing to pay to
  use the code under a commercial license).


\item {\it Trade secrets and competitive advantage:} Some authors feel
  that since their code is precious, by ``giving it away'' they give
  away their competitive advantage. If a ``competing'' lab needs six
  months to replicate my experiment, I can get further ahead in the
  meantime. While this position is tempting for the individual, we are
  seeking advantages in efficiency and quality for the entire
  community, including our taxpayer-supported funding
  agencies. Companies are under no obligation to serve the community,
  but they can get the benefits described above by first protecting
  their ideas with patents before publication. If groups withhold
  their code for their own interest and against the interest of the
  community, their work is manifestly less valuable than it could be,
  and should be evaluated accordingly. Conversely, releasing high
  quality code should enhance a group's reputation and success
  rates. This provides a feedback mechanism that reinforces code
  publication.
\end{enumerate}

\section{Encouraging Code Publication}

How can the publication of source code be made a community norm?
Assuming the existence of a few suitable protocols, how can
researchers be encouraged to use them? Though we believe the research
quality and efficiency benefits should persuade many researchers,
achieving such a large cultural change is likely to require activism
at various levels in the community. 

At the most executive level, organizations such as the IEEE could make
paper publication conditional on code publication, perhaps with
exceptions in extenuating circumstances. Such a policy seems
impossibly heavy-handed at the moment, though it might be possible for
individual journals and conferences. Perhaps a new journal or
conference could adopt this strategy as a differentiating feature: if
the arguments above are true, such a venue could expect to become
disproportionately influential. We cite some evidence of this effect
from other fields below.

  If {\it requiring} code publication seems
too ambitious, it is straightforward to {\it prefer} it. Publishers,
editors, program committees and individual reviewers can state that,
all else being equal, submissions that provide code are preferred over
those that do not. In practice, editors would need to advise reviewers
on the weighting of this preference, as with any other major criteria.

One simple concrete proposal is that the major conferences offer a new
prize for ``best'' (in quality, novelty or significance) published
code, along with the usual best paper and service prizes. This would
be a low cost, high visibility measure that recognizes this as a new
and significant way to contribute to the community.

At the most grassroots level, professors can expect their students to
back up all written work with published code. Generations of grad
students are short, and norms can be quickly established by
generational change.

Generational change may happen last to tenure and promotion
committees. We hope that academic departments will gradually come to
recognize code publication as a valuable academic activity. However,
we have argued above (and will provide evidence below) that papers
with code have more impact than papers without code. Better quality,
higher citation rates and good community visibility leading to stronger referee
letters are benefits that are already working under the traditional
evaluation criteria. 

We have argued above (Sections \ref{sec:eff1}, \ref{sec:eff2} points 1
\& 2) that increased code re-use can make work more efficient, so the
candidate's number of papers need not be reduced, but if this is a
concern, the following idea could help boost publication rates. When
an experiment is substantially re-used and the modifications reported,
the original author could be named as a co-author on the new
paper. This is not appropriate for middleware and simulation platform
code (e.g. Player and Stage), where normal citation is enough, but
rather when the code that embodies the idea of a specific experiment
is inherited. This is a way of rewarding production of impactful
experimental infrastructure, and is analagous to the long author lists
seen in for example genomics and astronomy, where infrastructure --
much of which is now software -- is recognized as extremely valuable.

\section{The Trend Toward Experiment Publishing}

We have argued that publishing code and experimental data is important
for the robotics research community, is good for researchers and easy
to do. Yet it is not standard practice in our field. The idea of
publishing experimental data and other artifacts beyond finished
papers is not new but it seems to be becoming popular. Here we survey
some government policies, practice in other
scientific disciplines and editorial policies of high impact journals.


\subsection{Government and Funding Agency Policies}

The US National Science Foundation
(NSF)... 

\begin{quotation}...expects investigators to share with other
  researchers, at no more than incremental cost and within a
  reasonable time, the data, samples, physical collections and other
  supporting materials created or gathered in the course of the
  work. It also encourages awardees to share software and inventions
  or otherwise act to make the innovations they embody widely useful
  and usable.\cite{nfs01grant} \end{quotation}

Since October 2003 the US National Institutes of Health (NIH) has
required grant applications for \$500K per year and above to include a
plan for data sharing or a statement why data sharing is not possible
\cite{nih03sharing}. While the form of data sharing is not considered
during the proposal assessment, the NIH sends a clear signal to
encourage publication of data.

The 2003 Berlin Declaration on Open Access to Knowledge
\cite{berlin03declaration} may come to be seen as an important
milestone. At the time of writing the declaration has been signed by
264 funding agencies, universities and research organizations,
including CERN, the Chinese Academy of Sciences, the Indian National
Science Academy, and the German Research Foundation.  The declaration
states:

\begin{quotation} A complete version of the work and all
  supplemental materials [...]
 in an appropriate standard electronic format is deposited
  (and thus published) in at least one online repository using
  suitable technical standards (such as the Open Archive definitions)
  that is supported and maintained by an academic institution,
  scholarly society, government agency, or other well established
  organization that seeks to enable open access, unrestricted
  distribution, inter operability, and long-term
  archiving. \cite{berlin03declaration}\end{quotation}

In a 2004 statement by the Organization for Economic Co-operation and
Development (OECD) numerous governments including those of North
America and Europe agreed on a declaration on access to research data
from public funding. The OECD recognizes that 
\begin{quotation}
an optimum international exchange of data, information and knowledge
contributes decisively to the advancement of scientific research and
innovation. [...] Open access to, and unrestricted use of, data
promotes scientific progress and facilitates the training of
researchers.  \cite{oecd04science}.\end{quotation}

In 2007 the US government passed the ``America COMPETES Act''
\cite{competes07} requiring federal civilian agencies that conduct
scientific research to openly exchange data and results with other
agencies, policymakers and the public.

All of these national and international governmental efforts are aimed
at improving the quality and efficiency of the science performed at
public expense, and each requires or requests that experimental data
and artifacts are shared.

\subsection{Practice in non-robotics disciplines}


According to Nielsen \cite{nielsen09doing} since 1991 physicists have
made extensive use of the preprint server \textit{arXiv}, which makes
papers freely available at the same time as they are submitted to a
journal for publication. Nielsen views \textit{arXiv} as an important
tool to speed up the transfer of knowledge, but goes further by
calling for the next generation of openness in science by ``... making
more types of content available than just scientific papers; allowing
creative reuse and modification of existing work through more open
licensing.''.

Arguably the life sciences are leading the trend. For example a US
National Academy of Sciences document on life science best practice
\cite{committe03sharing} requires authors to be consistent with the
principles of publication. This means that anything that is central to
a paper is to be made available in a way that enables replication,
verification and furtherance of science.  When it comes to publishing
algorithms, these guidelines are very explicit:

\begin{quotation}
...if the
intricacies of the algorithm make it difficult to describe in a
publication, the author could provide an outline of it in the paper
and make the source code [...] available to investigators...
\cite{committe03sharing}
\end{quotation}

In epidemiology usually more is at stake than in everyday
robotics. Epidemiological findings often influence policymakers, thus
society requires highly reliable results from this
field. Peng et al. \cite{peng06reproducible} acknowledge the sensitive
nature of this kind of work, and argue that \textit{reproducibility}
is the minimum standard for epidemiological research. Reproducibility
allows independent investigators to subject the original data to their
own analysis and interpretation. To enable reproducibility Peng
\begin{quotation}
...calls for data sets and software to be made available for 1)
verifying published findings, 2) conducting alternative analyzes of
the same data, 3) eliminating uninformed criticisms that do not stand
up to existing data, and 4) expediting the interchange of ideas among
investigators. \cite{peng06reproducible}
\end{quotation}

\subsection{Journals}

While scientists like Leonardo da Vinci, Galileo Galilei and
Christiaan Huygens kept their discoveries secret
\cite{nielsen09doing}, modern science is characterized by publication.
Britain's Royal Society mandated peer-review of published scientific
articles in its Philosophical Transactions, first published in
1665. This policy reflected the philosphy of the Royal Society,
expressed in their motto {\it Nullius in Verba} (nothing in words /
take nobody's word for it), that scientific claims are only valid
if reproducible. Since then, peer-reviewed journals have been the most
important way to communicate scientific results. Now, as the cost of
distributing large amounts of digital data becomes very low - a small
fraction of the total cost of an experiment, or of paper publication -
many journals require or at least encourage publication of data,
samples, code and detailed method descriptions alongside the
traditional paper.

\subsubsection{Science}
In 2004 and 2005 \textit{Science} published two stem cell papers
(\textit{Science} 303, 1669 (2004) and \textit{Science} 308, 1777
(2005)) which were later discovered to be fraudulent and retracted by the journal. As a consequence
\textit{Science} enlisted the help of an outside committee to
investigate the handling of the two papers and suggest improvements to
the editorial process of their journal. The committee concluded that
\textit{Science} had correctly followed their policies and that no
procedure could protect against deliberate fraud
\cite{kennedy06responding}. Interesting in the context of our paper is
the committee's recommendation for improving the editorial process
``\textit{Science} should have substantially stricter requirements
about reporting the primary data''.  Today \textit{Science} requires
that
\begin{itemize}
\item large data sets are deposited in approved public databases prior
  to publication and an accession number is included in the
  published paper.
\item all data necessary to understand, assess, and extend the
  conclusions of the paper must be made available to the reader,
  flowing the policies of \cite{nfs01grant} and
  \cite{committe03sharing}.
\item all reasonable requests for sharing materials are to be fulfilled.
 \cite{science09general} 
\end{itemize}

\subsubsection{Nature}

The Nature Publishing Group's policy on availability of data for all
their \textit{Nature} publications is very
similar:
\begin{quotation}An inherent principle of
publication is that others should be able to replicate and build upon
the authors' published claims. Therefore, a condition of publication
in a \textit{Nature} journal is that authors are required to make
material, data and associated protocols promptly available to readers
without preconditions.  \cite{nature09guide}\end{quotation} As with
\textit{Science}, \textit{Nature} requires depositing dataset in
publicly accessible databases. Of high interest in relation to our
paper is \textit{Nature's} policy on sharing biological materials. It
reads
\begin{quotation}
For materials such as mutant strains and cell lines, the \textit{Nature}
journals require authors to use established public repositories
whenever possible [...] and provide accession numbers in the manuscript. \cite{nature09guide}
\end{quotation}

Nature recently highlighted the importance of this issue on the cover
of an issue that featured a related editorial and three news/opinion
articles. The editorial described ``data's shameful neglect'', calling
for funding agencies to boost support for (and pressure on) researchers to
make data available \cite{nature:editorial}. Various reasons why many
researchers choose not to share, despite the existence of
purpose-built infrastructure, are discussed in the context of the
perceived failure of a digital archive project at the University of
Rochester\cite{nelson:nature}. A distinction is made between the
issues of pre-publication \cite{toronto:nature} and post-publication
\cite{schofield:nature} sharing of data and tools. We are advocating
{\it simultaneous} sharing and publication, which is a special case of
post-publication sharing. 

\subsubsection{Others}

Other journals like Nucleic Acids Research \cite{nar09general} and the
Public Library of Science journal series \cite{plos09editorial} have very
similar policies on data sharing and access to research material.

A less rigorous procedure is employed by the Annals of Internal
Medicine. To foster reproducible research and to enhance trust in
scientific results of publications, the journal encourages authors to
make their data publicly available and mandates authors to include a
statement of whether materials are being made available or not and if
under which conditions \cite{laime07reproducible}.

\subsubsection{Robotics}

The journal Autonomous Robots appears to have no stated policy on code
and data publication, though it does support the bundling of
supplementary material such as videos and data spreadsheets. The IEEE
Transactions on Robotics is similar, with no policy recommending data
or code sharing. However, it does mention these explicitly in the
``multimedia'' instructions:

\begin{quotation}
Multimedia can be ``playable'' files [...] or ``dataset'' files (e.g.,
raw data with programs to manipulate them). Such material is intended
to enhance the contents of a paper, both in clarity and in added
value.
\end{quotation}

\subsection{Impact on citation rates}

Citation counts are commonly used to assess the impact of an author's
work \cite{diamond86what}. A 2007 meta-study of cancer micorarray
clinical trials revealed that papers which shared their mircoarray
data were cited about 70\% more frequently \cite{piwowar07sharing}
than those that did not. If this effect generalizes to robotics,
authors would have an interest in publishing code in order to boost
their citation counts. An increase in citations can also be achieved
by publishing in open-access journals \cite{eysenbach06citation}.

\subsection{Related efforts}

A system for sharing and reproducing computations is proposed by
Schwab et al. \cite{schwab00making}, who use their system, based on
GNU {\it make}, as the principal means for organizing and transferring
scientific computations in their geophysics laboratory. The motivation
for ReDoc is essentially the same as that described in this
paper. However, perhaps unfortunately, this tool does not appear to
have made a large impact in computer science so far. ReDoc is clever
and powerful, but requires the user to learn to read special Make
macros. The method we advocate in this paper is more simple and does
not prescribe a particular build system, and can be thought of as a
subset of the ReDoc workflow.

\section{Sharing in Robotics}

Various free-to-use middleware and simulartion platforms for robotics
have been used over the last fifteen years, of which
Carmen\cite{carmen}, Player\cite{player2001} and recently
ROS\footnote{http://ros.org} are probably the most influential in research. Much of
the code-sharing that has occurred is facilitated by one of these
three platforms, each of which is the result of many hours of
work. 

ROS is currently the platform for a major
effort to bootstrap a community based around common
resources. In \label{sec:willow} 2010 Willow
Garage\footnote{http://willowgarage.com} used their unusual resources
to provide ten groups with highly capable PR2 mobile manipulator
robots running ROS software under the condition that resulting code be
made available for later users. While the PR2 is too expensive to be
ubiquitous, Willow Garage publishes a family of simulation models of
their robot based on well-known and free platforms. ROS goes much
further than Carmen and Player in encapsulating algorithms, including
recent SLAM, vision and control methods, and making them ready to use
as components of a downstream system. While ROS is open source and has
maby contributors, Willow Garage curates the project and does
significant development.

Willow Garage's expensive effort is the first to provide
state-of-the-art mechatronics and system engineering along with free
software tools, and is explicitly intended to accelerate research
progress by providing resources shared resources to the
community. While there will always be central place for custom robots
in research, more resources from components to complete systems will
become available in future. We can leverage the large investment in
open source software and projects like the PR2 by shifting to code
sharing as standard practice.

\section{Conclusion}

We have shown that meta-government organizations like the OECD see
scientific data exchange as an important tool for the efficient
advancement of scientific research. We have also argued that code is a
form of data that is particularly important for robotics
research. Thus making experimental data including code and
configurations publicly available is \textbf{important} for the
progress of robotics.

Building upon other peoples work is an integral part of the scientific
process. Increasing the efficiency of this process increases community
productivity. We suggest that making experimental code identifiable
will be helpful. We have also argued that the direct and indirect
effects of publishing identifiable code are \textbf{good} for the
researcher that shares, as well as for the wider community.

Other research fields, especially life sciences, have strong
requirements to share data sets and provide free access to supporting
materials alongside with traditional paper publications. In the case
of \textit{Nature} a submission of biological material to a public
repository may be required. In robotics, while sharing physical robots
may be prohibitively expensive, sharing digital resources takes little
time or treasure. Freely available infrastructure allows the upload of
a complete experiment (software, build scripts, data, analysis scripts
etc.) to a public repository in a few seconds with a few button
presses. Code sharing in robotics is \textbf{easy}.

The issues discussed here are not new, and a subset of robotics
researchers does publish source code implementations of their
algorithms, to the benefit of everyone. The main contribution here is
to point out the importance of distributing uniquely identifiable
versions and not just the latest and ``best'' version. We have argued
that this methodological issue is important, that originators and
subsequent users can both benefit, and suggested an easy-to-follow
publishing protocol. Our group will follow this protocol and observe
its effects.

\section*{Acknowledgements}

Thanks to Greg Mori, Alex Couture-Beil, Yaroslav Litus, Brian Gerkey,
Gaurav Sukhatme, and the organizers and attendees of the RSS'09
Workshop on Methodology in Experimental Robotics for useful
discussions on this issue.


\begin{thebibliography}{10}

\bibitem{knuth:literate}
D.E.~Knuth. 
\newblock Literate Programming.
\newblock {\em The Computer Journal} (British Computer Society), 27 (2) 97--111, 1984.

\bibitem{berlin03declaration}
Berlin declaration on open access to knowledge in sciences and humanities,
  2003.
\newblock \url{http://oa.mpg.de/openaccess-berlin/berlin_declaration.pdf}
  [Online; accessed 09-June 2009].

\bibitem{science09general}
Science: General information for authors, 2009.
\newblock
  \url{http://www.sciencemag.org/about/authors/prep/gen_info.dtl#datadep}
  [Online; accessed 10-June-2009].

\bibitem{committe03sharing}
T.~R. Cech et~al.
\newblock {\em Sharing Publication-Related Data and Materials: Responsibilities
  of Authorship in the Life Sciences}.
\newblock National Academy of Sciences, 2003.

\bibitem{diamond86what}
A.~M. {Diamond Jr.}
\newblock What is a citation worth.
\newblock {\em Journal of Human Resources}, 21(2):200--215, 1986.

\bibitem{eysenbach06citation}
G.~Eysenbach.
\newblock Citation advantage of open access articles.
\newblock {\em PLoS Biology}, 4(5):e157, 2006.

\bibitem{carmen}
M.~Montemerlo, N.~Roy, and S.~Thrun. 
\newblock Perspectives on standardization in mobile robot programming: The Carnegie Mellon navigation (CARMEN) toolkit. 
\newblock In {\em IEEE/RSJ Proceedings of the International Conference on Intelligent Robots and Systems},
 pages XX--YY, 2003. 

\bibitem{player2001}
B.~P. Gerkey, R.~T. Vaughan, K.~Stoy, A.~Howard, G.~S. Sukhatme, and M.~J.
  Matari\'{c}.
\newblock Most valuable player: A robot device server for distributed control.
\newblock In {\em IEEE/RSJ International Conference on Intelligent Robots and
  Systems}, pages 1226 -- 1231, 2001.

\bibitem{kennedy06responding}
D.~Kennedy.
\newblock Responging to fraud.
\newblock {\em Science}, 314(5804):1353, December 2006.

\bibitem{laime07reproducible}
C.~Laine, S.~N. Goodman, M.~E. Griswold, and H.~C. Sox.
\newblock Reproducible research: moving toward research the public can really
  trust.
\newblock {\em Annals of Internal Medicine}, 146(6):450–454, March 2007.

\bibitem{nih03sharing}
{National Institues of Health Office of Extramural Research}.
\newblock Nih data sharing policy, 2003.
\newblock \url{http://grants.nih.gov/grants/policy/data_sharing/} [Online;
  assessed 09-June-2009].

\bibitem{nfs01grant}
{National Science Foundation}.
\newblock Grant general conditions, 2001.
\newblock \url{http://www.nsf.gov/pubs/2001/gc101/gc101rev1.pdf} [Online;
  accessed 09-June-2009].

\bibitem{nature09guide}
Nature Editorial policies
\newblock Guide to publication policies of the nature journals, 2009.
\newblock \url{http://www.nature.com/authors/gta.pdf} [Online; accessed
  09-June-2009].

\bibitem{nature:editorial}
Nature Editorial
\newblock Data's shameful neglect.
\newblock {\em Nature}, 461(7271):145, 10 September 2009.

\bibitem{schofield:nature}
Schofield, P., Bubela, T., Weaver, T., Portilla, L., Brown, S.,
Hancock, J., Einhorn, D., Tocchini-Valentini, G., Hrabe de Angelis,
M., and Rosenthal, N. 
\newblock Post-publication sharing of data and tools.
\newblock {\em Nature}, 461(7271):171-173, 10 September 2009.

\bibitem{toronto:nature}
Toronto International Data Release Workshop Authors
\newblock Prepublication data sharing.
\newblock {\em Nature}, 461(7271):168-170, 10 September 2009.

\bibitem{nelson:nature}
Bryn Nelson
\newblock Data sharing: Empty archives
\newblock {\em Nature}, 461(7271):160-163, 10 September 2009.




\bibitem{nielsen09doing}
M.~Nielsen.
\newblock Doing science in the open.
\newblock {\em Physicsworld}, 22(5):30, 2009.

\bibitem{nar09general}
{Nucleic Acids Research}.
\newblock General policies of the journal, 2009.
\newblock
  \url{http://www.oxfordjournals.org/our_journals/nar/for_authors/ed_policy.ht%
ml} [Online; accessed 09-June-2009].

\bibitem{oecd04science}
{Organisation for Economic Co-operation and Development}.
\newblock Science, technology and innovation for the 21st. century. meeting of
  the {OECD} commitee for scientific and technological policy at ministerial
  level, January 2004.
\newblock
  \url{http://www.oecd.org/document/0,2340,en_2649_34487_25998799_1_1_1_1,00.h%
tml} [Online; accessed 09-June-2009].

\bibitem{peng06reproducible}
R.~D. Peng, F.~Dominici, and S.~L. Zeger.
\newblock Reproducible epidemiologic research.

\bibitem{piwowar07sharing}
H.~A. Piwowar, R.~S. Day, and D.~B. Fridsma.
\newblock Sharing detailed research data is associated with increased citation
  rate.
\newblock {\em PLoS ONE}, 2(3):e308, 2007.

\bibitem{plos09editorial}
{Public Library of Science (PLoS)}.
\newblock Editorial and publishing policies, 2009.
\newblock \url{http://www.plosone.org/static/policies.action} [Online; accessed
  10-June-2009].

\bibitem{RFC3174}
T.~Hansen and  G.~ Wollman
\newblock RFC3174: US Secure Hash Algorithm 1 (SHA1).
\newblock Technical report, Internet Engineering Task Force: Network Working
  Group, 2001.

\bibitem{schwab00making}
M.~Schwab, M.~Karrenbach, and J.~Clearbout.
\newblock Making scientific computations reproducible.
\newblock {\em Computing in Science and Engineering}, 2(6):61--67, 2000.

\bibitem{competes07}
{US Congress}.
\newblock {America COMPETES Act}, 2007.
\newblock \url{http://commdocs.house.gov/reports/110/h2272.pdf} [Online;
  accessed 09-June 2009].

\bibitem{WangMD5}
X.~Wang and H.~Yu.
\newblock How to break md5 and other hash functions.
\newblock In {\em Eurocrypt 2005}, volume 3494, pages 19--35. Lecture Notes in
  Computer Science, May 2005.

\end{thebibliography}
\bibliographystyle{abbrv}

\end{document}